\documentclass[11pt]{article}
\usepackage[margin=1in]{geometry}
\usepackage{amsmath,amsfonts}
\usepackage{algorithmic}
\usepackage{algorithm}
\usepackage{array}
\usepackage[caption=false,font=normalsize,labelfont=sf,textfont=sf]{subfig}
\usepackage{textcomp}
\usepackage{stfloats}
\usepackage{url}
\usepackage{verbatim}
\usepackage{graphicx}
\usepackage{booktabs}
\usepackage{multirow}
\usepackage{cite}
\usepackage{hyperref}
\begin{document}

\title{Reliability-Regulated Trajectory Optimization for Progressive COLMAP-Free 3D Gaussian Splatting}

\author{%
\begin{tabular}{c}
Zijian~Wu$^{1}$, Jinliang~Wang$^{2,3}$, Zidian~Lin$^{1}$, Ying~Song$^{1}$ \\
Ziqian~Lu$^{1}$, Hanjie~Ma$^{1}$, Zhen~Ye$^{4,5}$, and Mingfeng~Jiang$^{1,6}$ \\[0.75em]
\small $^1$School of Computer Science and Technology, Zhejiang Sci-Tech University \\
\small $^2$School of Biomedical Engineering, Tsinghua University \\
\small $^3$CardioCloud Medical Technology (Beijing) Co., Ltd. \\
\small $^4$Lishui University \\
\small $^5$State Key Laboratory of Blockchain and Data Security, Zhejiang University \\
\small $^6$College of Artificial Intelligence, Jiaxing University
\end{tabular}}
\date{}

\maketitle

\begin{abstract}
COLMAP-free 3D Gaussian Splatting (3DGS) bypasses computationally expensive structure-from-motion (SfM) pipelines, yet progressive camera pose tracking remains fundamentally vulnerable to error compounding---early pairwise tracking inaccuracies both corrupt subsequent frame initializations and remain permanently frozen in the scene representation. Rather than relying on heavyweight external neural priors or treating progressive tracking through isolated heuristic fixes, we propose a unified reliability-regulated trajectory optimization framework for progressive COLMAP-free 3DGS. At its core, our framework establishes an intrinsic, self-supervised bidirectional cycle-consistency mechanism that systematically regulates progressive camera trajectory estimation across two complementary temporal horizons:
(1) Forward Motion Propagation, where the online reliability signal adaptively gates first-order kinematic warm-starts of rigid motion into upcoming pairwise registrations, supplying informed directional search priors while safely intercepting untrusted transitions; and
(2) Retrospective Trajectory Correction, where the same reliability signal dynamically weights relative-pose consistency constraints within a sliding window of neighboring camera poses.
By governing both prospective state initialization and retrospective trajectory consolidation through a unified reliability regulator, our self-contained framework resolves progressive drift without external priors or offline preprocessing. Extensive evaluations on Tanks and Temples and CO3D-V2 benchmarks show that our method substantially improves camera trajectory accuracy and novel-view rendering quality, outperforming existing unposed baselines. Code is available at \url{https://github.com/Zijian1026/RRTO-CF3DGS}.
\end{abstract}


\section{Introduction}
High-fidelity 3D scene reconstruction and novel-view synthesis have experienced significant progress, largely driven by the evolution from implicit Neural Radiance Fields (NeRFs) \cite{mildenhall2021nerf} to explicit 3D Gaussian Splatting (3DGS) \cite{kerbl20233d}. While NeRFs model scenes through volumetric ray-marching via implicit neural networks, 3DGS introduces an explicit point-based scene representation parameterized by 3D Gaussians. Coupled with a fast, differentiable rasterization pipeline, 3DGS achieves real-time rendering speeds and faster optimization while preserving fine-grained geometric and photometric details. Consequently, 3DGS has rapidly established itself as a cornerstone paradigm for interactive 3D rendering and digital twin applications \cite{10870258,qian20243dgs,liu2024citygaussian,mildenhall2021nerf,szymanowicz2024splatter,11093421}.

Despite its efficiency, conventional 3DGS relies heavily on pre-computed, accurate camera poses, typically extracted via Structure-from-Motion (SfM) pipelines like COLMAP \cite{colmap}. In practice, obtaining reliable poses through SfM involves heavy computational overheads—such as feature matching and global bundle adjustment—and is highly sensitive to challenging conditions characterized by weak textures, repetitive patterns, or large-baseline camera movements \cite{huang20253r,li2024dngaussian,zhu2024fsgs,guo2025robust, ICLR2025_bf11e0bb}. Crucially, when SfM completely fails to establish feature correspondences, the downstream 3DGS pipeline cannot even be initialized. Even when SfM successfully yields sub-optimal or noisy poses, the subsequent Gaussian optimization easily degenerates, leading to severe visual artifacts and geometric distortion. Addressing this bottleneck requires pose-free reconstruction frameworks capable of jointly estimating scene geometry and camera motion directly from unposed image sequences \cite{wang2025vggt,fang2026dens3r}.

Earlier efforts to remove SfM constraints predominantly focused on NeRFs, such as BARF~\cite{barf} and Nope-NeRF~\cite{nope-nerf}, which jointly optimize neural fields and camera embeddings. However, the implicit nature of NeRFs renders pose optimization indirect and highly susceptible to local minima. Recently, CF-3DGS~\cite{cf-3dgs} marked a key milestone in this line of research. By leveraging the explicit point-cloud structure of 3DGS, CF-3DGS transforms camera estimation into direct rigid SE(3) transformations on 3D Gaussians. Through a progressive local-to-global sequential frame integration scheme, CF-3DGS not only eliminates the time-consuming SfM pre-processing step but also significantly outperforms previous NeRF-based pose-free methods \cite{nerfmm,SCNeRF2021} in both novel-view synthesis quality and trajectory estimation.

Nevertheless, the strictly sequential design of CF-3DGS leaves its trajectory optimization vulnerable to error accumulation. Because each newly registered frame relies transitively on previously estimated poses, localized relative transformation errors can gradually cascade over extended sequences, increasing the risk of unconstrained trajectory drift. This instability is further exacerbated under sparse observations or texture-scarce regions where photometric guidance alone provides insufficient optimization bounds. Furthermore, optimizing newly added frames independently fails to enforce temporal smoothness constraints among adjacent views, frequently causing physically implausible motion jitter even when localized image alignment appears photometrically sound. Crucially, a foundational vulnerability that remains unexplored in existing progressive 3DGS frameworks \cite{sun2024correspondence, ji2025sfm,wei2025pcr,chen2024zerogs,NEURIPS2025_7cdc4122} is the deceptive nature of pure photometric alignment during sequential pose registration. Through a deeper geometric analysis, we observe that localized photometric residuals often collapse into shallow local minima in textureless or repetitive regions, where incorrect camera poses can still render deceptive, low-error images by sacrificing structural sanity. When these noisy relative measurements are transitively committed, progressive pipelines inevitably suffer from a severe geometry-pose degeneracy coupling—where the 3D Gaussian geometry deforms to absorb camera tracking errors, making subsequent joint optimization incapable of recovering true trajectories.

\begin{figure}[t]
	\centering
	\includegraphics[width=\columnwidth]{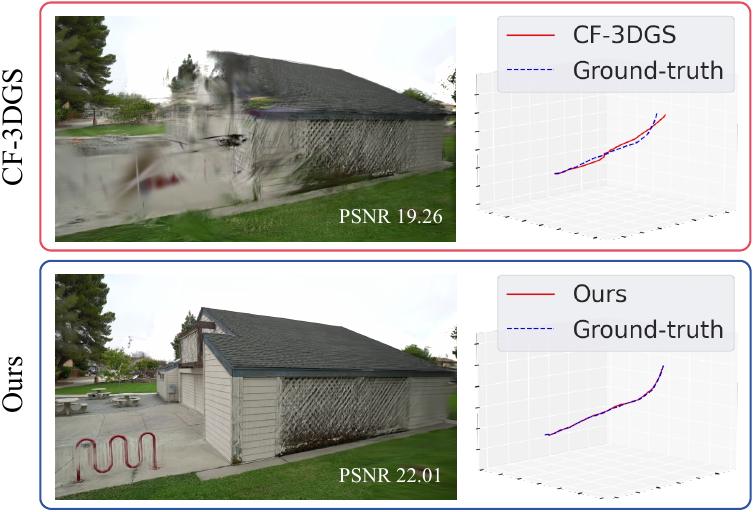}
    \caption{\textbf{Qualitative comparison on novel-view synthesis and camera trajectory estimation.} Our reliability-regulated trajectory optimization prevents local error compounding along the sequence and eliminates trajectory drift relative to CF-3DGS.}
	\label{FIG:1}
\end{figure}

To overcome these fundamental challenges without resorting to external priors \cite{fan2024instantsplat, venkatraman2026vggt}, we propose a unified reliability-regulated trajectory optimization framework for progressive COLMAP-free 3DGS. Rather than presenting fragmented heuristic modifications, our central insight is that progressive trajectory optimization must be systematically regulated across two distinct temporal horizons—prospective forward propagation and retrospective trajectory correction. Specifically, to regulate forward motion propagation and overcome initialization blindness without compounding errors along the sequence, we extrapolate the preceding relative transformation as a first-order kinematic warm-start for upcoming pairwise registrations, using our bidirectional cycle-consistency metric as a dynamic gate that intercepts untrusted transitions and falls back to identity. To regulate retrospective trajectory correction and eliminate accumulated drift while strictly avoiding geometry-pose degeneracy coupling, we freeze all 3D Gaussian parameters and periodically optimize neighboring camera poses within a sliding window. Poses are jointly constrained by multi-view photometric rasterization and a scale-normalized relative pose consistency term weighted by the same cycle-consistency reliability score.

Crucially, both horizons are governed by an online forward--backward cycle-consistency verification. Unlike single-view photometric residuals, which can remain misleadingly small in flat, textureless regions and produce false confidence, evaluating the dual-direction SE(3) cycle consistency between independent forward and backward Gaussian registrations yields a genuine geometric self-consistency proxy. This shared reliability score serves as an intrinsic regulator: it acts as a gate to prevent corrupted motion priors from contaminating new registrations, and acts as a dynamic loss weight that prevents noisy pairwise measurements from distorting retrospective window optimization.

Our framework preserves the progressive, real-time, and SfM-free merits of 3DGS while remaining entirely self-contained. Across extensive evaluations on Tanks and Temples and CO3D-V2 benchmarks, our approach yields smoother, highly accurate camera trajectories and delivers superior novel-view rendering quality. The main contributions are summarized as follows:
\begin{itemize}
    \item We identify and analyze the fundamental geometry-pose degeneracy coupling in progressive 3DGS, proposing a unified reliability-regulated trajectory optimization framework that resolves progressive drift without external neural priors.

    \item We introduce an intrinsic bidirectional cycle-consistency metric on $\mathrm{SE}(3)$ that acts as a unified geometric reliability regulator.

    \item We formulate a confidence-gated kinematic warm-start to govern forward motion propagation, alongside a fixed-geometry local window refinement scheme governed by confidence-weighted consistency to steer retrospective trajectory correction, effectively breaking the geometry-pose degeneracy loop.

    \item We demonstrate through extensive quantitative and qualitative evaluations that our framework substantially outperforms existing unposed baselines in both trajectory recovery accuracy and novel-view rendering fidelity.
\end{itemize}

\begin{figure*}[!t]
    \centering
    \includegraphics[width=\textwidth]{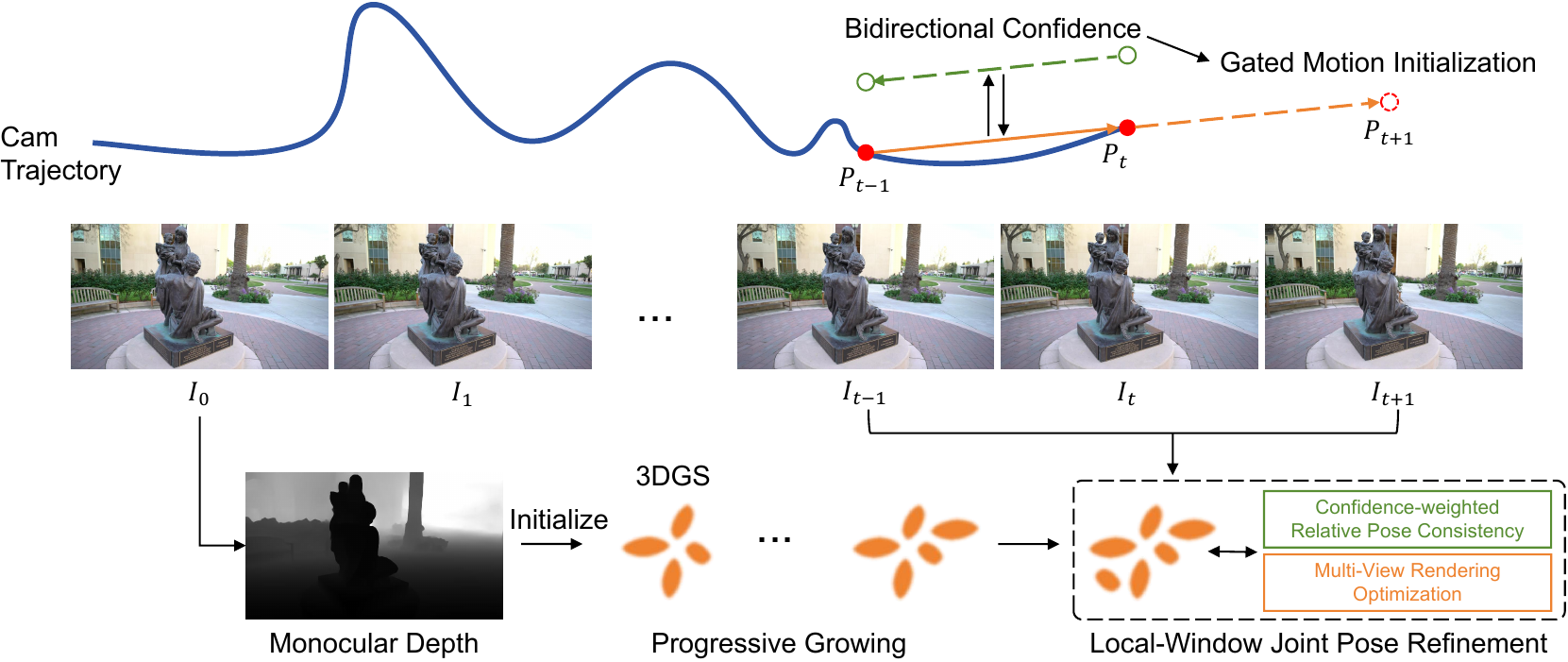}
    \caption{\textbf{Overview of the proposed reliability-regulated trajectory optimization framework.} Given an unposed image sequence, our framework establishes an intrinsic bidirectional cycle-consistency signal to systematically regulate progressive camera estimation across two complementary temporal phases: \textbf{(a) Forward Motion Propagation}, where tracking confidence adaptively gates first-order kinematic extrapolation on $\mathrm{SE}(3)$ to supply an informed warm-start while intercepting corrupted priors; and \textbf{(b) Retrospective Trajectory Correction}, where camera poses within a sliding window are jointly optimized under fixed Gaussian geometry, constrained by multi-view photometric alignment and confidence-weighted relative pose consistency. Shared cycle reliability ensures that noisy transitions neither corrupt future registrations nor distort local window trajectory refinement.}
    \label{fig:overview}
\end{figure*}

\section{Related Work}

\subsection{Novel View Synthesis}
Novel view synthesis aims to generate photorealistic images from unobserved camera viewpoints. NeRF~\cite{mildenhall2021nerf} revolutionized this field by parameterizing continuous 3D scenes using Multi-Layer Perceptrons (MLPs) optimized via volume rendering. While NeRF and its multi-scale variants~\cite{barron2021mip, barron2022mip,yu2021pixelnerf,Verbin_2022_CVPR,fang2025nerf, chen2021mvsnerf} achieve remarkable rendering fidelity, their implicit formulation requires dense ray-marching sampling, leading to extremely high computational overheads and slow rendering speeds.

To address these limitations, 3DGS~\cite{kerbl20233d} introduced an explicit point-based scene representation. By parameterizing scenes as a collection of learnable 3D Gaussians paired with a fast, tile-based differentiable rasterization pipeline, 3DGS achieves real-time rendering alongside significantly faster optimization times. Subsequent research has extended 3DGS across various domains, including scene compression \cite{c3dgs,mallick2024taming,ICLR2025_41e638f3}, anti-aliasing \cite{Yu2024MipSplatting,yu2024gsdf}, dynamic tracking \cite{4dgs,10757420, yang2023gs4d}, and large-scale urban reconstruction \cite{liu2024citygaussian, feng2025flashgs}. However, standard 3DGS remains tightly coupled with pre-computed camera poses and initial point clouds generated by SfM pipelines. When SfM fails or produces noisy poses, 3DGS optimization degrades severely, motivating the pursuit of pose-free paradigms.

\subsection{Pose-Free 3D Reconstruction and Neural Rendering}

Relaxing the requirement for pre-computed SfM camera poses has attracted significant interest in recent years. Early efforts primarily focused on jointly optimizing implicit NeRF representations and camera pose parameters without SfM preprocessing \cite{yang2023nerfvs,10.1145/3664647.3681507,wang2023f2nerf}. NeRFmm~\cite{nerfmm} introduced simultaneous optimization of NeRF parameters and camera pose embeddings, though it remains restricted to forward-facing scenes. BARF~\cite{barf} proposed a coarse-to-fine positional encoding mechanism to stabilize joint optimization, enabling robust pose alignment under complex motions provided a reasonable pose initialization. Nope-NeRF~\cite{nope-nerf} incorporated monocular depth priors to resolve scale ambiguity and relative pose drift during NeRF optimization. Nevertheless, optimizing poses within implicit representations relies indirectly on backpropagating gradients through implicit ray casting, leaving NeRF-based methods computationally expensive and prone to local minima under wide-baseline motions.

The explicit representation of 3DGS offers new opportunities for joint pose and geometry estimation, as $\mathrm{SE}(3)$ rigid transformations can be directly applied to 3D Gaussian centers. Existing pose-free 3DGS approaches target fundamentally distinct application scenarios. On the one hand, feed-forward generalizable models, such as FreeSplatter~\cite{Xu_2025_ICCV} and SPFSplat~\cite{Huang_2025_ICCV}, are specifically tailored for sparse-view 3D reconstruction from wide-baseline, unposed image pairs or triplets. By leveraging large transformer backbones pre-trained on massive datasets, they directly predict camera poses and 3D Gaussian primitives in a single pass \cite{Charatan_2024_CVPR}. While effective for rapid sparse-view interpolation, their application scope is strictly bounded by pre-training distributions and cannot handle continuous video streams or long sequential inputs. On the other hand, test-time progressive tracking frameworks are designed for continuous image sequences and video trajectories, performing test-time optimization without offline pre-training. CF-3DGS~\cite{cf-3dgs} marked a milestone in this sequential paradigm by incrementally estimating relative poses between adjacent frames using local 3DGS models to progressively expand global scene geometry. To enhance sequential stability, subsequent methods incorporate auxiliary learned priors; for instance, HT-3DGS~\cite{ji2025sfm} uses Video Frame Interpolation (VFI) neural networks to merge local segments, while PCR-GS~\cite{wei2025pcr} incorporates DINO \cite{oquab2023dinov2} feature reprojection and wavelet frequency constraints to regularize relative camera poses.

However, progressive tracking methods remain fundamentally vulnerable to cumulative trajectory drift over extended sequences, while recent attempts to mitigate this issue rely heavily on external neural models or multi-stage offline preprocessing \cite{cheng2025unposed3dgsreconstructionprobabilistic,fan2024instantsplat}. In contrast, our framework directly addresses the intrinsic trajectory estimation bottlenecks of progressive 3DGS in continuous sequence reconstruction. By exploiting first-order kinematic continuity for forward warm-starting and enforcing fixed-geometry local window multi-view consistency, unified by intrinsic forward--backward cycle consistency, our approach establishes a self-contained and drift-resilient progressive trajectory optimization pipeline without external priors or multi-stage training.

\section{Method}
Our framework establishes a unified trajectory optimization pipeline for progressive COLMAP-free 3DGS. As illustrated in Fig.~\ref{fig:overview}, the pipeline operates across two alternating phases, connected by a shared bidirectional tracking confidence that serves as a dynamic reliability signal. In the first phase, described in Sec.~\ref{sec:motion_prior}, short-term kinematic continuity on $\mathrm{SE}(3)$ is leveraged to provide an informed warm-start for newly added frames. This extrapolation is gated by the tracking confidence of the preceding transition to prevent untrusted motion priors from corrupting upcoming registration. In the second phase, detailed in Sec.~\ref{sec:window_refinement}, camera poses within a sliding temporal window are periodically unfrozen and jointly updated under fixed 3D Gaussian geometry. Within each window, poses are constrained by multi-view photometric alignment alongside a relative pose consistency loss weighted by the cached confidence scores. The following subsections detail our representation preliminaries, the bidirectional reliability evaluation, and how this shared signal systematically regulates both forward propagation and retrospective correction.
\subsection{Preliminaries}

\subsubsection{3D Gaussian Splatting}

3DGS~\cite{kerbl20233d} represents a scene as a
set of explicit 3D Gaussian primitives:
\begin{equation}
    \mathcal{G}_j =
    \left(
        \boldsymbol{\mu}_j,
        \boldsymbol{\Sigma}_j,
        \alpha_j,
        \mathbf{f}_j
    \right),
    \label{eq:gaussian_primitive}
\end{equation}
where $\boldsymbol{\mu}_j$, $\boldsymbol{\Sigma}_j$, $\alpha_j$, and
$\mathbf{f}_j$ denote the center, covariance, opacity, and appearance features
of the $j$-th Gaussian, respectively. The covariance is parameterized by a
rotation matrix $\mathbf{R}_j$ and a diagonal scaling matrix $\mathbf{S}_j$:
\begin{equation}
    \boldsymbol{\Sigma}_j =
    \mathbf{R}_j \mathbf{S}_j \mathbf{S}_j^{\mathsf{T}}
    \mathbf{R}_j^{\mathsf{T}}.
    \label{eq:gaussian_covariance}
\end{equation}

Given a camera pose, the Gaussians are projected onto the image plane and
rendered by differentiable alpha compositing:
\begin{equation}
    \widehat{\mathbf{C}}(\mathbf{p}) =
    \sum_{j=1}^{M}
    T_j(\mathbf{p})\alpha_j(\mathbf{p})\mathbf{c}_j,
    \label{eq:gaussian_rendering}
\end{equation}
where $\mathbf{c}_j$ is the projected color and $T_j(\mathbf{p})$ is the
accumulated transmittance of preceding Gaussians. Although this representation
supports efficient optimization and high-quality rendering, it is sensitive to
camera-pose errors, which can impair both geometric consistency and rendering
quality.

\subsubsection{COLMAP-Free 3D Gaussian Splatting}

Conventional 3DGS pipelines usually rely on camera poses obtained from a SfM system such as COLMAP. CF-3DGS~\cite{cf-3dgs} removes the dependence on externally estimated camera poses by progressively estimating camera motion and reconstructing the Gaussian scene. Given an image sequence
$\{I_k\}_{k=0}^{N-1}$, CF-3DGS first constructs a local Gaussian representation from a reference image using monocular depth prediction. The pose of a target image is then optimized by differentiable rendering against this local representation.

Let $P_k \in \mathrm{SE}(3)$ denote the world-to-camera transformation of
frame $k$. The first frame defines the initial coordinate system:
\begin{equation}
    P_0 = \mathbf{I}.
    \label{eq:initial_pose}
\end{equation}

For a newly observed frame $k$, CF-3DGS estimates the relative transformation
$\Delta_{k-1 \rightarrow k}$ between frames $k-1$ and $k$. The global camera
pose is updated by
\begin{equation}
    P_k =
    \Delta_{k-1 \rightarrow k} P_{k-1}.
    \label{eq:pose_composition}
\end{equation}

The relative transformation is estimated through differentiable rendering:
\begin{equation}
    \Delta_{k-1 \rightarrow k}^{*}
    =
    \arg\min_{\Delta \in \mathrm{SE}(3)}
    \mathcal{L}_{\mathrm{photo}}
    \left(
        \mathcal{R}
        \left(
            \mathcal{G}_{k-1}, \Delta
        \right),
        I_k
    \right),
    \label{eq:relative_pose_estimation}
\end{equation}
where $\mathcal{G}_{k-1}$ denotes the local Gaussian representation
constructed from the reference frame, $\mathcal{R}(\cdot)$ denotes the
differentiable rendering operation, and $\mathcal{L}_{\mathrm{photo}}$ denotes the
photometric reconstruction loss.

Once the camera pose $P_k$ is committed, CF-3DGS executes a frame replay step to update the Gaussian scene representation.

\subsection{Bidirectional Cycle-Consistency Evaluation}
\label{sec:forward_backward_verification}

In progressive pose estimation, tracking reliability can fluctuate significantly across frames due to texture scarcity, motion blur, or rapid perspective changes. To obtain an online self-supervised reliability proxy for each relative motion without ground-truth poses or external sensors, we perform bidirectional cycle-consistency verification.

Specifically, when registering an incoming target frame $k$ against its preceding reference frame $k-1$, forward differentiable tracking yields a predicted relative pose $\widehat{\Delta}_{k-1 \rightarrow k}$ using the depth-lifted local Gaussian model constructed from frame $k-1$. Immediately following forward registration, we independently estimate a backward relative transformation $\widehat{\Delta}_{k \rightarrow k-1}$ by tracking frame $k-1$ against the local Gaussian model constructed from frame $k$. Thus, the two directions are independently optimized renderer-based registrations rather than analytic inverses of the same estimate. For an ideal, geometrically consistent motion, their bidirectional composition should satisfy cycle consistency:
\begin{equation}
    C_{k-1,k}
    =
    \widehat{\Delta}_{k \rightarrow k-1} \widehat{\Delta}_{k-1 \rightarrow k}
    \approx \mathbf{I}.
    \label{eq:forward_backward_cycle}
\end{equation}

We evaluate the translational residual $e_t^k$ and rotational geodesic residual $e_R^k$ from the cycle transformation $C_{k-1,k}$:
\begin{align}
    e_t^k 
    &= \left\| \mathbf{t}(C_{k-1,k}) \right\|_2, \label{eq:forward_backward_trans_error} \\
    e_R^k 
    &= \arccos \left( \operatorname{clip} \left( \frac{\operatorname{tr}(\mathbf{R}(C_{k-1,k})) - 1}{2}, -1, 1 \right) \right). \label{eq:forward_backward_rot_error}
\end{align}
To eliminate scale dependency in monocular translation, we normalize $e_t^k$ by the forward translation magnitude:
\begin{equation}
    \widetilde{e}_t^k
    =
    \frac{e_t^k}{\max \left( \left\| \mathbf{t}(\widehat{\Delta}_{k-1 \rightarrow k}) \right\|_2, \epsilon \right)},
    \label{eq:normalized_forward_backward_error}
\end{equation}
where $\epsilon$ is a small positive constant ($\epsilon = 1.0 \times 10^{-3}$) that guarantees numerical stability and prevents denominator singularity during near-static frames or pure-rotational camera motions. Under pure rotational movements where translation magnitude approaches zero, the denominator is bounded by $\epsilon$, and the cycle metric is predominantly steered by the rotational residual $e_R^k$.

These bidirectional error metrics are mapped into a motion tracking confidence score $c_k \in (0, 1]$ via a smooth exponential decay formulation:
\begin{equation}
    c_k
    =
    \exp \left[ -\frac{1}{2} \left( \widetilde{e}_t^k + \frac{e_R^k}{\theta_0} \right) \right],
    \label{eq:forward_backward_confidence}
\end{equation}
where $\theta_0$ is the physically grounded scale constant for rotational deviation. The exponential mapping provides a smooth, bounded gradient that smoothly attenuates the influence of uncertain frames.

This confidence is computed on the fly during progressive registration and cached with its corresponding relative-pose measurement. It serves as an intrinsic, self-supervised reliability proxy that measures the dual-manifold reversibility of differentiable Gaussian tracking without requiring ground-truth poses. Importantly, forward--backward agreement is not treated as an absolute guarantee of global accuracy: in ambiguous or repetitive textures, mutually consistent yet biased tracking may occur. Hence, our framework does not replace photometric loss with confidence; rather, confidence acts as a safety gate for motion propagation (Sec.~\ref{sec:motion_prior}) and an adaptive soft anchor during window refinement (Sec.~\ref{sec:window_refinement}).

\subsection{Forward Motion Propagation with Kinematic Priors}
\label{sec:motion_prior}

Camera trajectories in real-world capturing possess short-term kinematic continuity, where adjacent frame-to-frame relative transformations exhibit smoothness in their Lie algebra velocity representation. However, standard progressive pipelines such as CF-3DGS initialize the pose optimization of every new frame with an identity transformation $\mathbf{I}$. Consequently, under texture-scarce or wide-baseline conditions, differentiable rasterization lacks a directional search prior, rendering it highly prone to local photometric minima.

To provide an informed optimization trajectory, we formulate a kinematic motion warm-start strategy. Assuming local first-order continuity in the camera's relative motion tangent space $\mathfrak{se}(3)$, the relative transformation from frame $k-2$ to $k-1$ serves as a first-order extrapolation prior for the incoming transformation from $k-1$ to $k$:
\begin{equation}
    \boldsymbol{\xi}_{k-1 \rightarrow k}^{(0)} \approx \boldsymbol{\xi}_{k-2 \rightarrow k-1} \implies \Delta_{k-1 \rightarrow k}^{(0)} = \widehat{\Delta}_{k-2 \rightarrow k-1}.
\end{equation}
However, indiscriminately applying motion extrapolation is perilous: when the camera abruptly changes direction or when the preceding tracking is corrupted, extrapolating a bad estimate compounds trajectory errors catastrophically. To eliminate this vulnerability, we gate the motion extrapolation using the cached forward--backward confidence $c_{k-1}$:
\begin{equation}
    \Delta_{k-1 \rightarrow k}^{(0)}
    =
    \begin{cases}
        \widehat{\Delta}_{k-2 \rightarrow k-1}, & \text{if } c_{k-1} \ge \tau_c, \\
        \mathbf{I}, & \text{otherwise},
    \end{cases}
    \label{eq:motion_prior}
\end{equation}
where $\tau_c$ is a preset confidence threshold. When the preceding transition exhibits verified dual-direction geometric consistency ($c_{k-1} \ge \tau_c$), the kinematic extrapolation is adopted as an informed warm start; if tracking is deemed untrusted ($c_{k-1} < \tau_c$), the initialization safely falls back to the zero-bias identity transformation $\mathbf{I}$. This design ensures that motion extrapolation accelerates and guides convergence in well-behaved segments while remaining fail-safe against abrupt trajectory perturbations.

\subsection{Retrospective Trajectory Correction via Joint Refinement}
\label{sec:window_refinement}

Existing progressive COLMAP-free 3DGS methods typically adopt a strictly sequential ``commit-and-freeze'' paradigm, where each camera pose is frozen permanently after its initial pairwise registration. Under this scheme, localized tracking errors accumulate unchecked across frames, precipitating severe trajectory drift.

A naive adaptation of classical Bundle Adjustment (jointly optimizing 3D Gaussians and camera poses) fails fundamentally due to the extreme geometric plasticity of 3DGS. Because 3D Gaussians can freely alter their positions, scales, and opacities, jointly updating scene geometry and camera parameters allows the Gaussians to deform and absorb camera pose errors, permanently baking trajectory drift into distorted scene geometry. 

To overcome this, we establish a fixed-geometry local window joint refinement scheme. We periodically unfreeze a temporal sliding window of consecutive camera poses $\mathcal{W} = \{s, s+1, \ldots, e\}$ while strictly freezing all 3D Gaussian scene parameters. By fixing scene geometry, error gradients derived from multi-view differentiable rendering are forced to flow entirely into camera pose parameters $\boldsymbol{\xi}_i \in \mathfrak{se}(3)$, breaking the geometry-pose degeneracy loop.

The sliding window operates with a size of $|\mathcal{W}|$ frames, advancing by one frame upon each newly registered image. Consequently, each camera pose is iteratively co-optimized across multiple overlapping window configurations, achieving temporal consensus. Within each window, the anchor pose $P_s$ is fixed to establish a stable reference gauge, while the remaining poses $\{P_i\}_{i=s+1}^e$ are jointly refined using two complementary objectives: multi-view photometric alignment and scale-normalized, confidence-weighted relative pose consistency.

\begin{figure*}[t]
    \centering
    \includegraphics[width=\textwidth]{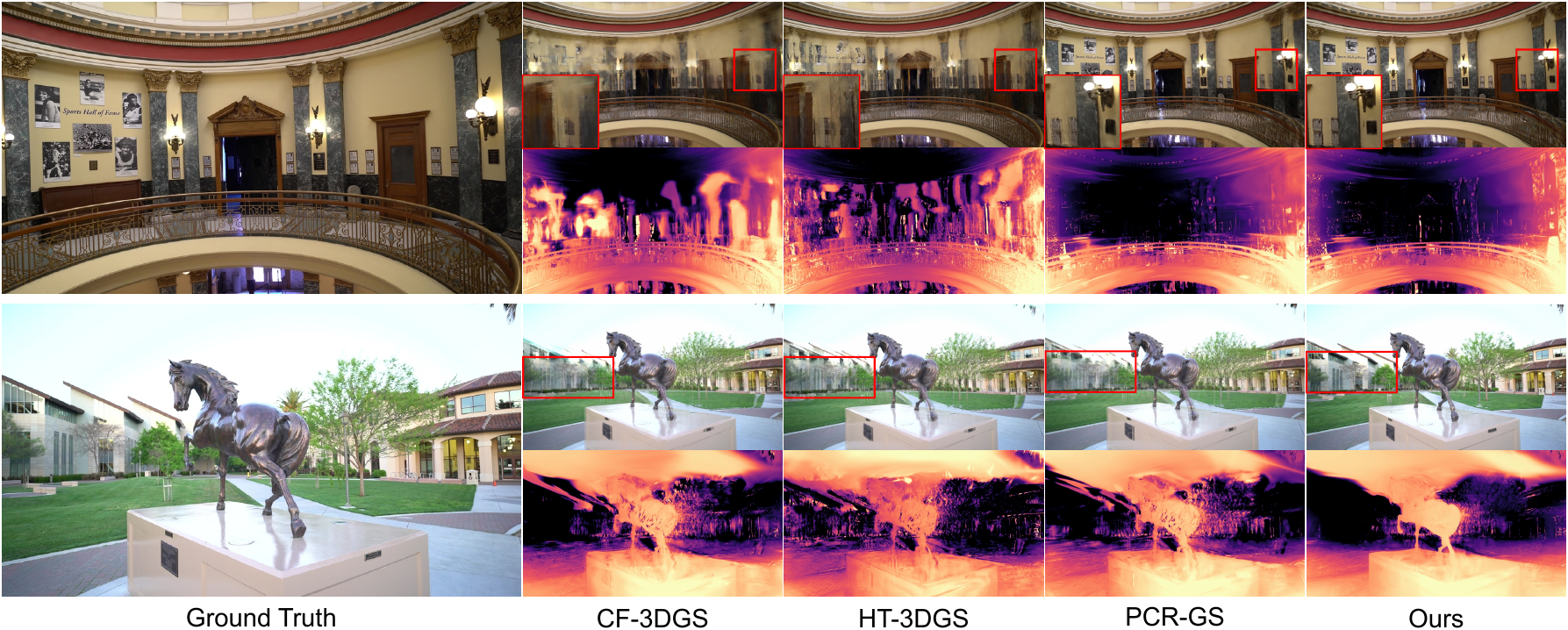}
    \caption{\textbf{Qualitative novel-view synthesis results on the Tanks and Temples benchmark.} We compare our method against unposed baselines. While existing methods exhibit severe blurring, missing structures, and geometric distortion due to noisy or drifted pose estimates, our approach produces photorealistic renderings with fine-grained textures and sharp edges. Inset red and green boxes highlight key detailed regions.}
    \label{fig:tanks_qualitative}
\end{figure*}
\begin{table*}[ht]
\centering
\small
\caption{Quantitative novel view synthesis results on the Tanks and Temples benchmark with stride 5. Held-out test views are evaluated using PSNR~$\uparrow$, SSIM~$\uparrow$, and LPIPS~$\downarrow$. Best results are bolded.}
\label{tab:tanks_nvs}
\setlength{\tabcolsep}{3pt}
\resizebox{\textwidth}{!}{%
\begin{tabular}{l|ccc|ccc|ccc|ccc|ccc}
\toprule
\multirow{2}{*}{Scene} & \multicolumn{3}{c|}{Nope-NeRF~\cite{nope-nerf}} & \multicolumn{3}{c|}{CF-3DGS~\cite{cf-3dgs}} & \multicolumn{3}{c|}{HT-3DGS~\cite{ji2025sfm}} & \multicolumn{3}{c|}{PCR-GS~\cite{wei2025pcr}} & \multicolumn{3}{c}{\textbf{Ours}} \\
& PSNR & SSIM & LPIPS & PSNR & SSIM & LPIPS & PSNR & SSIM & LPIPS & PSNR & SSIM & LPIPS & PSNR & SSIM & LPIPS \\
\midrule
Ballroom & 24.43 & 0.71 & 0.28 & 24.92 & 0.83 & 0.12 & 25.62 & 0.85 & 0.11 & 25.16 & 0.83 & 0.12 & \textbf{28.16} & \textbf{0.91} & \textbf{0.07} \\
Barn     & 23.00 & 0.61 & 0.40 & 19.26 & 0.57 & 0.38 & 20.33 & 0.56 & 0.35 & 19.00 & 0.53 & 0.46 & \textbf{22.01} & \textbf{0.67} & \textbf{0.27} \\
Church   & 23.01 & 0.65 & 0.37 & 26.83 & 0.87 & 0.14 & \textbf{27.80} & 0.89 & 0.12 & 27.31 & 0.89 & 0.13 & 27.53 & \textbf{0.89} & \textbf{0.12} \\
Family   & 24.59 & 0.71 & 0.35 & 25.03 & 0.85 & 0.16 & 27.80 & 0.89 & 0.12 & 27.15 & 0.90 & 0.12 & \textbf{28.31} & \textbf{0.91} & \textbf{0.11} \\
Francis  & 23.01 & 0.66 & 0.43 & 23.90 & 0.74 & 0.27 & 24.31 & 0.72 & 0.26 & 24.77 & 0.73 & 0.26 & \textbf{25.96} & \textbf{0.77} & \textbf{0.25} \\
Horse    & 19.53 & 0.67 & 0.36 & 17.56 & 0.60 & 0.28 & 17.81 & 0.61 & 0.28 & 17.86 & 0.61 & 0.28 & \textbf{21.84} & \textbf{0.72} & \textbf{0.21} \\
Ignatius & 21.68 & 0.53 & 0.45 & 21.76 & 0.69 & 0.21 & \textbf{22.47} & \textbf{0.77} & \textbf{0.16} & 21.98 & 0.70 & 0.20 & 21.82 & 0.69 & 0.21 \\
Museum   & 21.39 & 0.65 & 0.34 & 16.19 & 0.52 & 0.44 & 17.48 & 0.54 & 0.42 & 18.40 & 0.50 & 0.39 & \textbf{23.09} & \textbf{0.73} & \textbf{0.22} \\
\midrule
Mean     & 22.580 & 0.649 & 0.373 & 21.931 & 0.709 & 0.250 & 22.953 & 0.729 & 0.228 & 22.704 & 0.711 & 0.245 & \textbf{24.840} & \textbf{0.786} & \textbf{0.183} \\
\bottomrule
\end{tabular}
}
\end{table*}
\subsubsection{Photometric Joint Alignment}

Rather than optimizing each view independently against a single reference frame, we enforce local multi-view geometric consistency by jointly rendering all frames in the window against the current Gaussian representation. The window-level photometric alignment objective is formulated as:
\begin{equation}
    \mathcal{L}_{\mathrm{photo}}^{\mathrm{window}}
    =
    \frac{1}{|\mathcal{W}|}
    \sum_{i \in \mathcal{W}}
    \mathcal{L}_{\mathrm{photo}}
    \left(
        \mathcal{R}(\mathcal{G}, P_i),
        I_i
    \right),
    \label{eq:window_photo_loss}
\end{equation}
where $\mathcal{R}(\mathcal{G}, P_i)$ denotes the differentiable rasterization of the current Gaussian model $\mathcal{G}$ under camera pose $P_i$. Joint optimization over multi-view image observations supplies local photometric evidence that can resolve ambiguities not addressed by a single reference view. The relative-pose consistency term defined next uses the reliability information associated with earlier sequential tracking measurements.

\begin{figure*}[t]
    \centering
    \includegraphics[width=\textwidth]{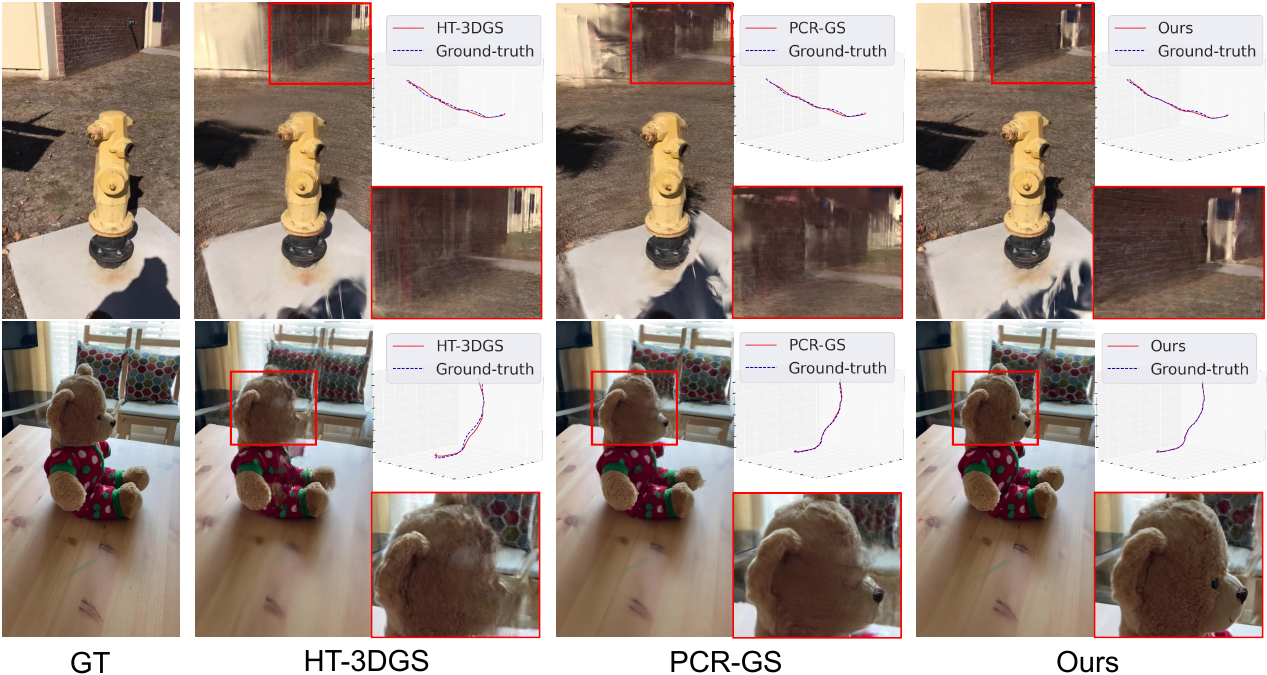}
    \caption{\textbf{Qualitative comparison on challenging 360-degree object-centric sequences from CO3D-V2.} On the displayed sequences with wide baselines and continuous rotational trajectories, competing methods exhibit trajectory deviations and rendering artifacts such as floaters and blurriness. Our reliability-regulated framework produces more accurate trajectory estimates and higher-fidelity Gaussian reconstructions.}
    \label{fig:co3d_qualitative}
\end{figure*}

\subsubsection{Confidence-Weighted Relative-Pose Consistency}

Photometric optimization alone can be under-constrained in texture-scarce regions or wide-baseline frame transitions. To stabilize the proposed local window joint refinement, we introduce a scale-normalized, confidence-weighted relative-pose consistency term. Each cached relative motion is treated as a reliability-aware soft measurement during local refinement rather than as a uniform pose-smoothness constraint: its influence is inherited from the bidirectional verification performed when that exact motion was estimated.

Let $\Delta_{i-1 \rightarrow i}^{\mathrm{pred}} = P_i P_{i-1}^{-1}$ denote the relative pose implied by the current window trajectory. We penalize deviations between $\Delta_{i-1 \rightarrow i}^{\mathrm{pred}}$ and the initial tracked measurement $\widehat{\Delta}_{i-1 \rightarrow i}$ using a confidence-weighted loss:
\begin{equation}
    \mathcal{L}_{\mathrm{cons}}
    =
    \frac{\sum_{i=s+1}^{e} c_i \left( \mathcal{L}_{R}^i + \mathcal{L}_{t}^i \right)}{\max\left(\sum_{i=s+1}^{e} c_i, \epsilon\right)},
    \label{eq:confidence_weighted_consistency}
\end{equation}
where $\mathcal{L}_{R}^i$ penalizes rotational discrepancy and $\mathcal{L}_{t}^i$ penalizes scale-normalized translational discrepancy. To ensure numerical stability and avoid gradient singularities near identity rotations, the rotational loss $\mathcal{L}_R^i$ is defined as the element-wise mean squared error:
\begin{equation}
    \mathcal{L}_R^i = \operatorname{mean}\left[ \left( \mathbf{R}_{i-1 \rightarrow i}^{\mathrm{pred}} \left(\widehat{\mathbf{R}}_{i-1 \rightarrow i}\right)^{\mathsf{T}} - \mathbf{I} \right)^2 \right],
    \label{eq:rot_loss_frobenius}
\end{equation}
and the translational loss is formulated with a scale-normalized Smooth-$L_1$ objective:
\begin{equation}
    \begin{aligned}
        \mathcal{L}_t^i &= \operatorname{Smooth}_{L_1}\left(\frac{\mathbf{t}_{i-1 \rightarrow i}^{\mathrm{pred}}}{s_t^i}, \frac{\widehat{\mathbf{t}}_{i-1 \rightarrow i}}{s_t^i}\right), \\
        \text{where} \quad s_t^i &= \max\left(\left\|\widehat{\mathbf{t}}_{i-1 \rightarrow i}\right\|_2, \epsilon\right).
    \end{aligned}
    \label{eq:trans_loss_smoothl1}
\end{equation}

\begin{table}[t]
\centering
\small
\caption{Average novel view synthesis results on the CO3D-V2 dataset. Test views are evaluated using PSNR~$\uparrow$, SSIM~$\uparrow$, and LPIPS~$\downarrow$. Best results are bolded.}
\label{tab:co3d_nvs}
\setlength{\tabcolsep}{8pt}
\begin{tabular}{l|ccc}
\toprule
Method & PSNR~$\uparrow$ & SSIM~$\uparrow$ & LPIPS~$\downarrow$ \\
\midrule
CF-3DGS~\cite{cf-3dgs}       & 26.11 & 0.782 & 0.270 \\
HT-3DGS~\cite{ji2025sfm} & 26.59 & 0.800 & 0.260 \\
PCR-GS~\cite{wei2025pcr}          & 26.16 & 0.785 & 0.272 \\
\textbf{Ours}                     & \textbf{27.71} & \textbf{0.819} & \textbf{0.239} \\
\bottomrule
\end{tabular}
\end{table}

The normalization by $s_t^i$ scales the translational residual using the magnitude of its cached relative-motion measurement, avoiding dependence on its absolute monocular translation scale. By anchoring the window-refined trajectory to high-confidence tracking measurements while downweighting low-consistency transitions (small $c_i$), this term preserves reliable temporal evidence without forcing ambiguous measurements to dominate the local photometric objective. Thus, the same forward--backward reliability score that gates motion-guided initialization also determines the strength of the corresponding constraint during local trajectory correction.

\subsection{Overall Optimization Objectives}
\label{sec:overall_objectives}

Our framework harmonizes forward motion propagation and retrospective trajectory correction through the following objectives. Crucially, the confidence $c_i$ used in the retrospective window loss is the exact same reliability score that regulates forward kinematic propagation in Eq.~\ref{eq:motion_prior}, embodying a cohesive, dual-horizon trajectory regulation mechanism.

\paragraph{Photometric Loss.}
Given a rendered view $\widehat{I}$ and target ground-truth view $I$, the photometric loss $\mathcal{L}_{\mathrm{photo}}$ combines $L_1$ and D-SSIM reconstruction terms:
\begin{equation}
    \mathcal{L}_{\mathrm{photo}}(\widehat{I}, I) = (1 - \lambda_{\mathrm{dssim}})\mathcal{L}_1(\widehat{I}, I) + \lambda_{\mathrm{dssim}}\mathcal{L}_{\mathrm{D-SSIM}}(\widehat{I}, I).
    \label{eq:photometric_loss_def}
\end{equation}
During the progressive registration phase, $\mathcal{L}_{\mathrm{photo}}$ drives both local relative pose tracking and the subsequent sequential update of 3D Gaussian attributes.

\paragraph{Window Refinement Loss.}
During the retrospective window refinement phase, the camera poses within window $\mathcal{W}$ are jointly refined by minimizing the multi-task window loss:
\begin{equation}
    \mathcal{L}_{\mathrm{window}} = \mathcal{L}_{\mathrm{photo}}^{\mathrm{window}} + \lambda_{\mathrm{cons}} \mathcal{L}_{\mathrm{cons}},
    \label{eq:window_total_loss}
\end{equation}
Together with Eq.~\ref{eq:motion_prior}, this objective realizes our unified reliability regulation: gating prospective motion propagation before a new pose is estimated and weighting retrospective trajectory correction when neighboring poses are revisited.

\begin{table*}[ht]
\centering
\small
\caption{Quantitative camera pose estimation results on the Tanks and Temples benchmark with stride 5. Trajectories are evaluated using $\text{RPE}_t$~$\downarrow$ ($\times 10^2$), $\text{RPE}_r$~$\downarrow$ ($^\circ$), and $\text{ATE}$~$\downarrow$. Best results are bolded.}
\label{tab:tanks_pose}
\setlength{\tabcolsep}{4.2pt}
\resizebox{\textwidth}{!}{%
\begin{tabular}{l|ccc|ccc|ccc|ccc|ccc}
\toprule
\multirow{2}{*}{Scene} & \multicolumn{3}{c|}{Nope-NeRF~\cite{nope-nerf}} & \multicolumn{3}{c|}{CF-3DGS~\cite{cf-3dgs}} & \multicolumn{3}{c|}{HT-3DGS~\cite{ji2025sfm}} & \multicolumn{3}{c|}{PCR-GS~\cite{wei2025pcr}} & \multicolumn{3}{c}{\textbf{Ours}} \\
& $\text{RPE}_t$ & $\text{RPE}_r$ & $\text{ATE}$ & $\text{RPE}_t$ & $\text{RPE}_r$ & $\text{ATE}$ & $\text{RPE}_t$ & $\text{RPE}_r$ & $\text{ATE}$ & $\text{RPE}_t$ & $\text{RPE}_r$ & $\text{ATE}$ & $\text{RPE}_t$ & $\text{RPE}_r$ & $\text{ATE}$ \\
\midrule
Ballroom & 0.127 & 0.247 & \textbf{0.002} & 0.279 & 0.048 & 0.006 & 0.279 & 0.047 & 0.006 & 0.295 & 0.059 & 0.006 & \textbf{0.122} & \textbf{0.026} & \textbf{0.002} \\
Barn     & 1.343 & 0.403 & 0.033 & 2.409 & 0.585 & 0.065 & 2.252 & 0.574 & 0.057 & 3.379 & 1.074 & 0.118 & \textbf{0.576} & \textbf{0.314} & \textbf{0.011} \\
Church   & 0.490 & 0.127 & 0.044 & 0.081 & 0.095 & 0.004 & 0.081 & 0.095 & 0.004 & 0.070 & 0.092 & 0.005 & \textbf{0.053} & \textbf{0.087} & \textbf{0.003} \\
Family   & \textbf{0.119} & \textbf{0.052} & \textbf{0.003} & 0.775 & 0.294 & 0.011 & 0.203 & 0.118 & 0.004 & 0.483 & 0.187 & 0.006 & 0.455 & 0.181 & 0.006 \\
Francis  & 2.941 & 1.515 & 0.079 & 0.814 & 0.916 & 0.021 & 0.733 & 0.878 & 0.016 & 0.714 & 0.816 & 0.015 & \textbf{0.416} & \textbf{0.747} & \textbf{0.015} \\
Horse    & 6.588 & 1.537 & 0.028 & 2.152 & 0.686 & 0.024 & 2.183 & 0.691 & 0.024 & 2.312 & 0.709 & 0.027 & \textbf{1.229} & \textbf{0.415} & \textbf{0.024} \\
Ignatius & \textbf{0.099} & \textbf{0.026} & \textbf{0.002} & 0.290 & 0.123 & 0.009 & 0.289 & 0.122 & 0.009 & 0.289 & 0.120 & 0.009 & 0.287 & 0.129 & 0.008 \\
Museum   & 0.741 & 1.134 & 0.026 & 3.814 & 2.809 & 0.055 & 3.809 & 2.807 & 0.055 & 3.979 & 1.709 & 0.139 & \textbf{0.735} & \textbf{0.873} & \textbf{0.009} \\
\midrule
Mean     & 1.556 & 0.630 & 0.027 & 1.327 & 0.695 & 0.024 & 1.229 & 0.667 & 0.022 & 1.440 & 0.596 & 0.041 & \textbf{0.484} & \textbf{0.347} & \textbf{0.010} \\
\bottomrule
\end{tabular}
}
\end{table*}

\begin{table}[ht]
\centering
\small
\caption{Average camera pose estimation results on the CO3D-V2 dataset. Trajectories are evaluated using $\text{RPE}_t$~$\downarrow$ ($\times 10^2$), $\text{RPE}_r$~$\downarrow$ ($^\circ$), and $\text{ATE}$~$\downarrow$. Best results are bolded.}
\label{tab:co3d_pose}
\setlength{\tabcolsep}{8pt}
\begin{tabular}{l|ccc}
\toprule
Method & $\text{RPE}_t$~$\downarrow$ & $\text{RPE}_r$~$\downarrow$ & $\text{ATE}$~$\downarrow$ \\
\midrule
CF-3DGS~\cite{cf-3dgs}       & 0.579 & 0.548 & 0.023 \\
HT-3DGS~\cite{ji2025sfm} & 0.545 & 0.511 & 0.022 \\
PCR-GS~\cite{wei2025pcr}          & 0.658 & 0.842 & 0.032 \\
\textbf{Ours}                     & \textbf{0.352} & \textbf{0.382} & \textbf{0.014} \\
\bottomrule
\end{tabular}
\end{table}

\section{Experiments}
\label{sec:experiments}

\subsection{Experimental Setup}
\label{sec:experimental_setup}

\subsubsection{Datasets}
We evaluate our method on two challenging benchmarks: Tanks and Temples~\cite{tanks} and CO3D-V2~\cite{co3d}. For Tanks and Temples, in order to better simulate the drastic camera
motions, we subsampling frames with a stride of 5. For CO3D-V2, we evaluate on circular object-centric sequences across six diverse categories (\textit{apple}, \textit{ball}, \textit{book}, \textit{hydrant}, \textit{suitcase}, and \textit{teddybear}). In both benchmarks, every eighth frame is reserved as a held-out view for testing, while the remaining frames are used for training.
\subsubsection{Baselines}
We benchmark our approach against representative state-of-the-art pose-free baselines: the coordinate-based implicit model Nope-NeRF~\cite{nope-nerf}, the original sequential CF-3DGS baseline~\cite{cf-3dgs}, the multi-level training framework HT-3DGS~\cite{ji2025sfm}, and the co-regularized variant PCR-GS~\cite{wei2025pcr}.

\subsubsection{Evaluation Metrics}
We evaluate both camera pose estimation and novel view synthesis. For camera trajectory estimation, we report the Absolute Trajectory Error ($\text{ATE}$) and the Relative Pose Errors for translation ($\text{RPE}_t$, $\times 10^2$) and rotation ($\text{RPE}_r$, in degrees) \cite{barf, nope-nerf}. For novel view synthesis on held-out test views, we measure visual quality using standard metrics: Peak Signal-to-Noise Ratio (PSNR), Structural Similarity (SSIM)~\cite{ssim}, and Learned Perceptual Image Patch Similarity (LPIPS)~\cite{lpips}.

\subsubsection{Implementation Details}
Our method is implemented in PyTorch using the 3DGS rasterization pipeline. Camera poses are parameterized as Lie algebra elements $\boldsymbol{\xi} \in \mathfrak{se}(3)$ and optimized with Adam at a learning rate of $1.0 \times 10^{-3}$. When registering a new frame, the local reference Gaussian model is fitted for $1{,}000$ iterations, followed by $300$ iterations each for forward and backward relative pose tracking. In the cycle-consistency verification that defines the shared forward--backward confidence, we set $\theta_0 = 5^\circ$, $\epsilon = 1.0 \times 10^{-3}$, and use a motion-initialization confidence threshold of $\tau_c = 0.85$. The relative-pose consistency weight is set to $\lambda_{\mathrm{cons}} = 0.1$, and the photometric loss uses $\lambda_{\mathrm{dssim}} = 0.2$. The sliding window has a size of $|\mathcal{W}| = 5$ frames with an overlap of $4$ frames (stride of 1). Local window joint pose refinement is optimized for $N_{\mathrm{joint}} = 100$ iterations on CO3D-V2 ($30$ on Tanks and Temples).

\subsection{Novel View Synthesis Evaluation}
\label{sec:nvs_eval}

We first evaluate novel view synthesis quality on held-out test viewpoints across both Tanks and Temples (Table~\ref{tab:tanks_nvs}) and CO3D-V2 (Table~\ref{tab:co3d_nvs}).

\begin{figure*}[t]
\centering
\includegraphics[width=\textwidth]{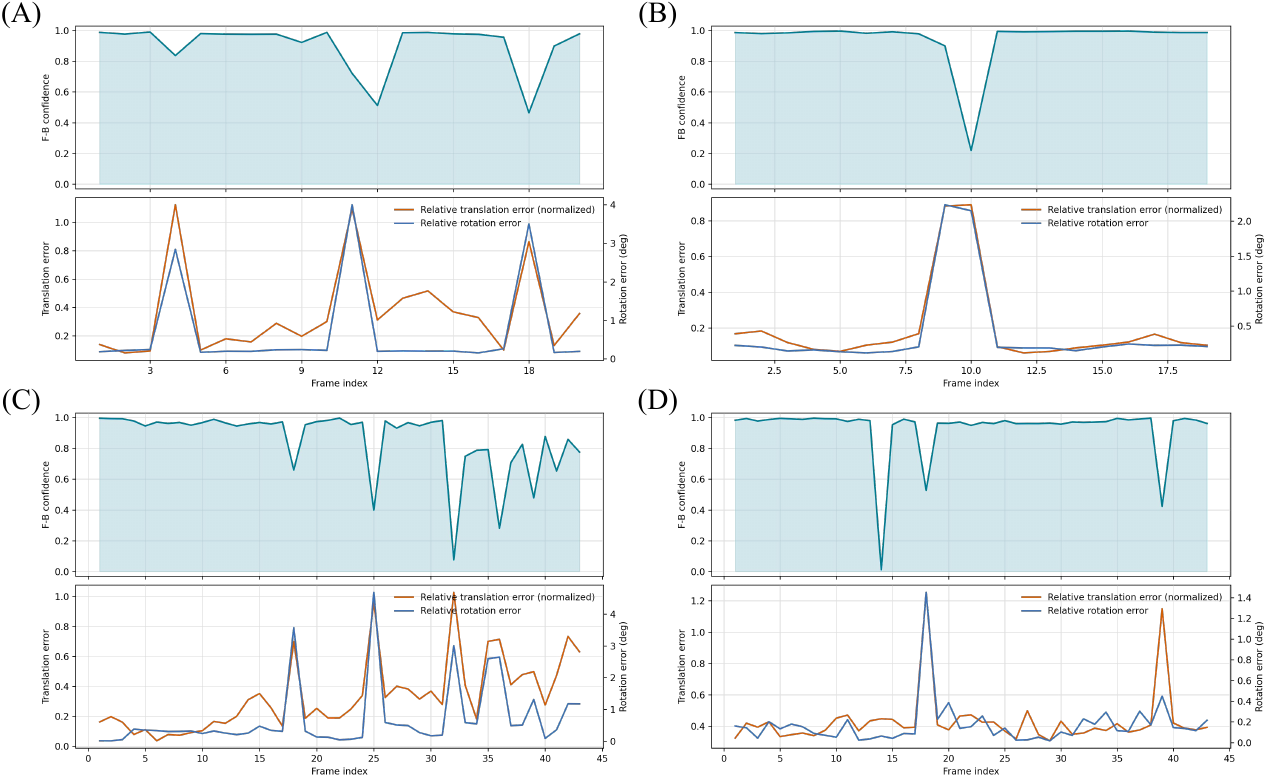}
\caption{\textbf{Reliability verification of bidirectional cycle-consistency confidence.} We visualize the temporal trajectory error against the estimated cycle-consistency confidence $c_i$ across representative sequences from Tanks and Temples (\textbf{A}: \textit{Family}, \textbf{B}: \textit{Horse}) and CO3D-V2 (\textbf{C}: \textit{Apple}, \textbf{D}: \textit{Book}). For each sequence, the upper subplot shows the frame-wise bidirectional confidence $c_i$, while the lower subplot displays the corresponding translational error and rotational discrepancy.}
\label{fig:confidence_verification}
\end{figure*}

\begin{table*}[ht]
\centering
\scriptsize
\caption{\textbf{Ablation study and reliability calibration on Tanks and Temples.} We evaluate the impact of motion-guided initialization (MP), confidence gating (Gate), window joint refinement (WJ), and forward--backward (F--B) consistency weighting. $\text{RPE}_t$ is reported in $\times 10^2$ and $\text{RPE}_r$ in degrees.}
\label{tab:ablation_study}
\setlength{\tabcolsep}{4pt}
\renewcommand{\arraystretch}{1.08}
\resizebox{\textwidth}{!}{%
\begin{tabular}{lcccl|ccc|ccc}
\toprule
\multirow{2}{*}{Variant} & \multicolumn{3}{c}{Controls} & \multirow{2}{*}{Loss Weight} & \multicolumn{3}{c|}{Rendering quality} & \multicolumn{3}{c}{Pose accuracy} \\
& MP & Gate & WJ & & PSNR$\uparrow$ & SSIM$\uparrow$ & LPIPS$\downarrow$ & $\text{RPE}_t\downarrow$ & $\text{RPE}_r\downarrow$ & $\text{ATE}\downarrow$ \\
\midrule
CF-3DGS (sequential) & $\times$ & -- & $\times$ & -- & 21.93 & 0.709 & 0.250 & 1.327 & 0.695 & 0.024 \\
WJ, photometric only & $\times$ & -- & $\checkmark$ & -- & 22.56 & 0.724 & 0.247 & 1.047 & 0.623 & 0.022 \\
WJ + Uniform loss weight & $\times$ & -- & $\checkmark$ & Uniform & 22.74 & 0.726 & 0.242 & 0.998 & 0.618 & 0.021 \\
WJ + F--B loss weight & $\times$ & -- & $\checkmark$ & F--B & 22.97 & 0.729 & 0.237 & 0.920 & 0.610 & 0.019 \\
\midrule
Ungated MP & $\checkmark$ & $\times$ & $\times$ & -- & 23.58 & 0.752 & 0.210 & 0.714 & 0.476 & 0.016 \\
Confidence-gated MP & $\checkmark$ & $\checkmark$ & $\times$ & -- & 24.26 & 0.780 & 0.187 & 0.513 & 0.360 & 0.011 \\
\midrule
Full model (Uniform weighted) & $\checkmark$ & $\checkmark$ & $\checkmark$ & Uniform & 24.51 & 0.781 & 0.187 & 0.502 & 0.355 & 0.011 \\
Full model (F--B weighted) & $\checkmark$ & $\checkmark$ & $\checkmark$ & F--B & \textbf{24.84} & \textbf{0.786} & \textbf{0.183} & \textbf{0.484} & \textbf{0.347} & \textbf{0.010} \\
\bottomrule
\end{tabular}
}

\end{table*}

On Tanks and Temples, our approach achieves substantial average visual-fidelity gains. Compared with CF-3DGS, it improves mean PSNR from $21.93$~dB to $24.84$~dB ($+2.91$~dB), raises SSIM from $0.709$ to $0.786$, and reduces LPIPS from $0.250$ to $0.183$. It also exceeds HT-3DGS and PCR-GS by more than $1.88$~dB in mean PSNR, with particularly large margins on \textit{Museum} ($+4.69$~dB over PCR-GS) and \textit{Horse} ($+3.98$~dB over PCR-GS). Figure~\ref{fig:tanks_qualitative} illustrates corresponding improvements in detail preservation and structural coherence on representative scenes. On CO3D-V2, the proposed method obtains the highest average rendering quality among the compared methods, reaching $27.71$~dB PSNR and $0.819$ SSIM. Together with the camera trajectory estimation results, these substantial visual fidelity gains directly validate the synergistic design of our two-horizon trajectory optimization framework.

\subsection{Camera Pose Estimation Evaluation}
\label{sec:pose_eval}

We next evaluate camera pose recovery accuracy on Tanks and Temples (Table~\ref{tab:tanks_pose}) and CO3D-V2 (Table~\ref{tab:co3d_pose}). Consistent with the evaluation protocol, estimated trajectories are aligned with ground-truth camera poses via $\mathrm{Sim}(3)$ Umeyama alignment, and evaluated using $\text{RPE}_t$, $\text{RPE}_r$, and $\text{ATE}$.

As shown in Table~\ref{tab:tanks_pose}, our method substantially reduces average trajectory estimation errors on Tanks and Temples. It achieves a mean $\text{ATE}$ of $0.010$, compared with $0.024$ for CF-3DGS, $0.022$ for HT-3DGS, $0.027$ for Nope-NeRF, and $0.041$ for PCR-GS. It also reduces mean $\text{RPE}_t$ to $0.484$ (a $63.5\%$ reduction relative to CF-3DGS) and mean $\text{RPE}_r$ to $0.347^\circ$ (a $50.0\%$ reduction relative to CF-3DGS). The largest improvements occur on \textit{Barn} and \textit{Museum}, where our reliability-regulated framework maintains $\text{RPE}_t$ below $0.74$ while CF-3DGS and PCR-GS exceed $2.4$.

On CO3D-V2 (Table~\ref{tab:co3d_pose}), where consecutive frames involve rapid circular rotations around compact objects, our method obtains the lowest average errors: mean $\text{RPE}_t$ of $0.352$, $\text{RPE}_r$ of $0.382^\circ$, and $\text{ATE}$ of $0.014$. These values correspond to reductions of $39.2\%$ in $\text{RPE}_t$, $30.3\%$ in $\text{RPE}_r$, and $39.1\%$ in $\text{ATE}$ relative to CF-3DGS, supporting the benefit of our reliability-regulated optimization on the evaluated object-centric trajectories.

\subsubsection{Reliability Verification of Bidirectional Confidence}

To empirically validate whether our forward--backward cycle-consistency confidence $c_i$ serves as a faithful, online proxy for pairwise tracking reliability, we visualize the temporal alignment between estimated confidence scores and true camera pose errors across representative sequences in Fig.~\ref{fig:confidence_verification}. 

As demonstrated across sequences from both Tanks and Temples and CO3D-V2, the estimated cycle confidence exhibits a strong inverse correlation with tracking inaccuracy. Specifically, sharp drops in confidence $c_i$ consistently coincide with local error spikes in translation and rotation. This strong temporal alignment confirms that cycle-consistency evaluation successfully isolates unreliable motion estimates without requiring ground-truth trajectory supervision.

We note that slight conservatism can occasionally occur in edge cases; for instance, around frame 14 of the \textit{Book} sequence (Fig.~\ref{fig:confidence_verification}D), the confidence drops despite a relatively low instantaneous pose error. Rather than a flaw, this conservative behavior acts as a desirable safety filter: when frame-to-frame geometric alignment is under-constrained, tracking consistency cannot be rigorously guaranteed, making it safer to fall back to identity initialization than to risk propagating potentially corrupted motion priors. Overall, across the vast majority of frame transitions, our bidirectional confidence acts as an exceptionally accurate and dependable ranker of real-time pose tracking quality.

\subsection{Ablation Study}
\label{sec:ablation_analysis}

To thoroughly dissect the working mechanisms and validate the empirical designs of our framework, we conduct extensive ablation experiments on the Tanks and Temples benchmark. Table~\ref{tab:ablation_study} reports both novel view synthesis and camera pose estimation metrics across carefully controlled baseline variants, where MP denotes the Motion-Guided Initialization prior, Gate represents our confidence-gated fallback mechanism, and WJ signifies local window joint pose refinement.

\paragraph{Efficacy of Kinematic Warm-Start and Confidence Gating.}
Comparing sequential tracking against motion-guided configurations demonstrates the critical necessity of an informed search prior. Naive sequential CF-3DGS yields only $21.93$\,dB PSNR and $0.024$ ATE due to initialization blindness. Incorporating ungated motion extrapolation (Ungated motion propagation) dramatically improves tracking convergence ($23.58$\,dB PSNR, $0.016$ ATE). However, blindly propagating preceding motion causes overshoot when the camera alters direction or when previous tracking was noisy. Adding our forward--backward confidence gate (Confidence-gated motion propagation) effectively filters out these degenerate priors, elevating PSNR to $24.26$\,dB and reducing ATE to $0.011$. This confirms that confidence gating is essential to prevent error compounding along physical trajectories.

\paragraph{Necessity of Confidence-Weighted Relative Pose Regularization.}
Within local window joint refinement, relying exclusively on multi-view photometric alignment (WJ, photometric only) provides modest improvements over standard CF-3DGS ($22.56$\,dB PSNR, $0.022$ ATE), because multi-view photometric rasterization remains ill-posed in textureless or low-parallax viewpoints. Enforcing a uniform relative pose constraint (WJ + Uniform loss weight) mitigates unconstrained drift ($22.74$\,dB PSNR, $0.021$ ATE). Weighting these relative-pose constraints by our bidirectional cycle consistency (WJ + F--B loss weight) further anchors the optimization against photometric ambiguities, reducing ATE to $0.019$. Crucially, when integrated into the full pipeline, our dynamic F--B loss weighting (Full model, F--B weighted) outperforms unweighted equal loss weighting (Full model, Uniform weighted) by $+0.33$\,dB in PSNR ($24.84$\,dB vs. $24.51$\,dB) and achieves the overall lowest trajectory error ($0.010$ ATE, $0.484$ $\text{RPE}_t$). This confirms that F--B confidence weighting successfully downweights noisy pairwise measurements during joint window pose refinement.

\section{Conclusion}

In this work, we presented a unified reliability-regulated trajectory optimization framework for progressive COLMAP-free 3D Gaussian Splatting. Rather than applying disconnected heuristic additions or relying on external learned priors, our framework establishes an intrinsic bidirectional cycle-consistency mechanism that systematically regulates camera trajectory estimation across two complementary temporal horizons: (1) forward motion propagation, where tracking reliability adaptively gates first-order kinematic warm-starts into upcoming frame registrations to avoid initialization blindness without compounding errors; and (2) retrospective trajectory correction, where the same reliability signal dynamically weights relative-pose consistency during local window optimization under fixed Gaussian geometry, strictly preventing geometry-pose degeneracy coupling. By tying prospective state initialization and retrospective trajectory refinement through a shared, self-supervised reliability regulator, our approach prevents local error accumulation and eliminates trajectory drift in sequential pose-free reconstruction. Extensive evaluations on challenging real-world benchmarks, including Tanks and Temples and CO3D-V2, demonstrate that our method significantly improves camera trajectory accuracy and novel-view rendering quality, outperforming existing unposed NeRF and 3DGS baselines.

\bibliographystyle{unsrt}
\bibliography{refs}

\end{document}